\documentclass[sigconf]{acmart}
\usepackage[utf8]{inputenc}
\usepackage{longtable}
\usepackage{diagbox}

\def\BibTeX{{\rm B\kern-.05em{\sc i\kern-.025em b}\kern-.08emT\kern-.1667em\lower.7ex\hbox{E}\kern-.125emX}}
    
\copyrightyear{2019}
\acmYear{2019}
\setcopyright{acmlicensed}
\acmConference[BDAI 2019]{The 2nd International Conference on Big Data and Artificial}{June 22-24, 2019}{Guangzhou, China}
\acmPrice{}
\acmDOI{}
\acmISBN{}

\begin{document}

%
\title[Semi-Unsupervised Lifelong Learning for Sentiment Classification]{Semi-Unsupervised Lifelong Learning for Sentiment Classification: Less Manual Data Annotation and More Self-Studying}

%
\author{Xianbin Hong}
\email{Xianbin.Hong@liverpool.ac.uk}
\orcid{0000-0003-1678-0948}
\affiliation{%
  \institution{Research Institute of Big Data Analytics, Xi'an Jiaotong-Liverpool University}
  \streetaddress{111, Ren Ai Road}
  \city{Suzhou}
  \state{Jiangsu}
  \country{China}
  \postcode{215123}
}

\author{Gautam Pal}
\email{Gautam.Pal@xjtlu.edu.cn}
\affiliation{%
  \institution{Research Institute of Big Data Analytics, Xi'an Jiaotong-Liverpool University}
  \streetaddress{111, Ren Ai Road}
  \city{Suzhou}
  \state{Jiangsu}
  \country{China}
  \postcode{215123}
}

\author{Sheng-Uei Guan}
\email{Steven.Guan@xjtlu.edu.cn}
\affiliation{%
  \institution{Research Institute of Big Data Analytics, Xi'an Jiaotong-Liverpool University}
  \streetaddress{111, Ren Ai Road}
  \city{Suzhou}
  \state{Jiangsu}
  \country{China}
  \postcode{215123}
}

\author{Prudence Wong}
\email{PWong@liverpool.ac.uk}
\affiliation{%
  \institution{Department of Computer Science, The University of Liverpool }
  \streetaddress{Ashton Building, Ashton Street}
  \city{Liverpool}
  \state{Merseyside}
  \country{UK}
  \postcode{L69 3BX}
}
 
\author{Dawei Liu}
\email{Dawei.Liu@xjtlu.edu.cn}
\affiliation{%
  \institution{Research Institute of Big Data Analytics, Xi'an Jiaotong-Liverpool University}
  \streetaddress{111, Ren Ai Road}
  \city{Suzhou}
  \state{Jiangsu}
  \country{China}
  \postcode{215123}
}

\author{Ka Lok Man}
\email{Ka.Man@xjtlu.edu.cn}
\affiliation{%
  \institution{Research Institute of Big Data Analytics, Xi'an Jiaotong-Liverpool University}
  \streetaddress{111, Ren Ai Road}
  \city{Suzhou}
  \state{Jiangsu}
  \country{China}
  \postcode{215123}
}

\author{Xin Huang}
\email{Xin.Huang@xjtlu.edu.cn}
\affiliation{%
  \institution{Research Institute of Big Data Analytics, Xi'an Jiaotong-Liverpool University}
  \streetaddress{111, Ren Ai Road}
  \city{Suzhou}
  \state{Jiangsu}
  \country{China}
  \postcode{215123}
}

%
\renewcommand{\shortauthors}{Xianbin Hong, et al.}

%
\begin{abstract}
Lifelong machine learning is a novel machine learning paradigm which can continually accumulate knowledge during learning. The knowledge extracting and reusing abilities enable the lifelong machine learning to solve the related problems. The traditional approaches like Na\"ive Bayes and some neural network based approaches only aim to achieve the best performance upon a single task. Unlike them, the lifelong machine learning in this paper focus on how to accumulate knowledge during learning and leverage them for the further tasks. Meanwhile, the demand for labeled data for training also be significantly decreased with the knowledge reusing. This paper suggests that the aim of the lifelong learning is to use less labeled data and computational cost to achieve the performance as well as or even better than the supervised learning.
\end{abstract}

%
%


%

%
\maketitle

\section{Introduction}
Over the past 30 years, machine learning have achieved a significant development. However, we are still in a era of "Weak AI" rather than "Strong AI". Current machine learning algorithms only know how to solve a specific problem but have no idea when they meet some related problems. Hence, the lifelong machine learning (simply said as lifelong learning or "LML" below) \cite{thrun1998lifelong} was raised to solve a infinite sequence of related tasks by knowledge accumulation and reusing. For the related problems, an integrated model with knowledge reusing could decrease the cost for the sample annotation. 

For instance, in the sentiment classification we need to predict the sentiment (positive or negative) of a sentence or a document. For different sentiment classification tasks, traditional approaches need to train an independent model on each domain to obtain the best performance. Hence, for each domain we need to collect labeled data for the supervised learning. In this way, the algorithm will never know how to solve a problem without new labeled data. This is what a typical "weak AI".

To achieve the goal of "strong AI", we need to change our learning goal to really understand the sentiment of words. Which means that the algorithm should know how each word influences the sentiment of a document in different tasks. If we can achieve this learning goal, the algorithms are able to solve new tasks without teaching. Zhiyuan Chen and etc. \cite{chen2015lifelong} ever proposed a approach to close the goal. They made a big progress but the supervised learning still is needed. Guangyi Lv and etc. \cite{lv2019sentiment} extend the work of \cite{chen2015lifelong} with a neural network based approach. However, the supervised learning still is necessary under their setting and huge volume of labeled data are required. Hence, this paper aims to decrease the usage of labeled data while maintain the performance.

\section{Lifelong Machine Learning}

It was firstly called as lifelong machine learning since 1995 by
Thrun \cite{thrun1995lifelongRobot,thrun1996learningn-ththing}. Efficient Lifelong Machine Learning (ELLA) \cite{ruvolo2013ella} raised by Ruvolo and Eaton. Comparing with the multi-task learning \cite{caruana1997multitask}, ELLA is much more efficient. Zhiyuan and etc. \cite{chen2015lifelong} improved the sentiment classification by involving knowledge. The object function was modified with two penalty terms which corresponding with previous tasks.

\subsection{Components of LML}

\begin{figure}[h]
  \centering
  \includegraphics[width=\linewidth]{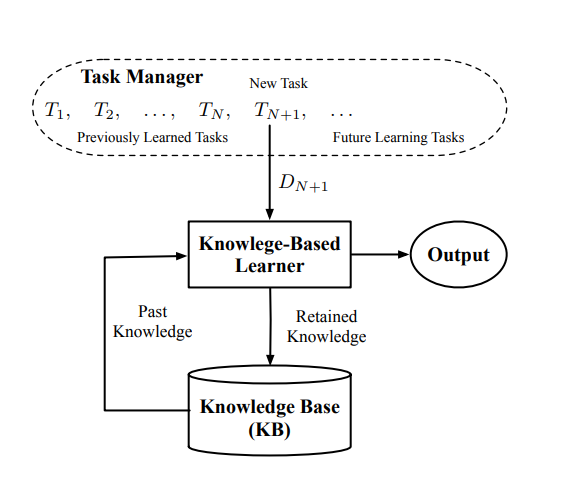}
  \caption{Knowledge System in the Lifelong Machine Learning \cite{chen2015lifelong}}
\end{figure}

The knowledge system contains the following components:
\begin{itemize}
\item {Knowledge Base (KB)}: The knowledge Base\cite{chen2015lifelong} mainly used to maintain the previous knowledge. Based on the type of knowledge, it could be divided as Past Information Store (PIS), Meta-Knowledge Miner (MKM) and Meta-Knowledge Store (MKS).
\item {Knowledge Reasoner (KR)}: The knowledge reasoner is designed to generate new knowledge upon the archived knowledge by logic inference. A strict logic design is required so the most of the LML algorithms lack of the component.
\item {Knowledge-Base Learner (KBL)}: The Knowledge-Based Learner\cite{chen2015lifelong} aims to retrieve and transfer previous knowledge to the current task. Hence, it contains two parts: task knowledge miner and leaner. The miner seeks and determines which knowledge could be reused, and the learner transfers such knowledge to the current task.
\end{itemize}

\subsection{Sentiment Classification}
Hong and etc.\cite{hong2018lifelong} had discussed that the NLP field is most suitable for the lifelong machine learning researches due to its knowledge is easy to extract and to be understood by human. Previous classical paper\cite{chen2015lifelong} chose the sentiment classification as the learning target because it could be regarded as a large task as well as a group of related sub-tasks in the different domains. Although these sub-tasks are related to each other but a model only trained on a single sub-tasks is unable to perform well in the rest sub-tasks. This requires the algorithms could know when the knowledge can be used and when can not due to the distribution of each sub-tasks is different. Known these, an algorithm can be called as "lifelong" because it is able to transfer previous knowledge to new tasks to improve performance. 

Although deep learning already is applied in sentiment classification, it still could not leverage past knowledge well. This because the complexity of neural network limits the researches to define and extract knowledge from the data. As the previous work\cite{chen2015lifelong},  this paper also uses Na\"ive Bayes as the knowledge can be presented by the probability. In this way, we need to know the probability of each word that shows in the positive or negative content. We also need to know well that some words may only have sentiment polarity in some specific domains(equal to tasks in this paper). "Lifelong Sentiment Classification" ("LSC" for simple below) \cite{chen2015lifelong} records that which domain does a word have the sentiment orientation. If a word always has sentiment polarity or has significant polarity in current domain, a higher weight will sign to it more than other words. This approach contains a knowledge transfer operation and a knowledge validation operation.

\section{Contribution of This Paper}

Although LSC\cite{chen2015lifelong} already raised a lifelong approach, it only aims to improve the classification accuracy. It still is under the setting of the supervised learning and also is unable to deliver an explicit knowledge to guild further learning.

Based on the LSC, this paper advances the lifelong learning in sentiment classification and have two main contributions:
\begin{itemize}
\item{\bf A improved lifelong learning paradigm is proposed to solve the sentiment classification problem under unsupervised learning setting with previous knowledge.}
\item{\bf We introduce a novel approach to discover and store the words with sentiment polarity for reuse.}
\end{itemize}

\section{Sentiment Polarity Words}
\subsection{Na\"ive Bayesian Text Classification}

In this paper, we define a word has sentiment polarity by calculating the probability that it appears in a positive or negative content (sentence or document). If a word has a high probability with sentiment polarity, it also will leads to the document have higher probability of sentiment probability based on the Na\"ive Bayesian (NB) formula. Hence, to determine the words with polarity is the key to predict the sentiment.

Na\"ive Bayesian (NB) classifier \cite{mccallum1999text} calculates the probability of each word $w$ in a document $d$ and then to predict the sentiment polarity (positive or negative). We use the same formula below as in the LSC\cite{chen2015lifelong}. $P(w|c_{j})$ is the probability of a word appears in a class:

\begin{equation}
P(w|c_{j}) = \frac{\lambda +N_{c_{j},w}}{\lambda \left | V \right | + \sum_{v=1}^{V}N_{c_{j},v}}    
\end{equation}

Where $c_{j}$ is either positive (+) or negative (-) sentiment polarity.  $N_{c_{j},w}$ is the frequency of a word w in documents of class $c_{j}$ . |V| is the size of vocabulary V and $\lambda (0 \leqslant \lambda \leqslant 1)$ is used for smoothing ( set as 1 for Laplace smoothing in this paper).

Given a document, we can calculate the probability of it for different classes by:

\begin{equation}
    P(c_{j}|d_{i}) = \frac{P(c_{j})\prod_{w\in d_{i}}P(w|c_{j})^{n_{w},d_{i}}}{\sum_{r=1}^{C} P(c_{r}) \prod_{w\in d_{i}}P(w|c_{r})^{n_{w},d_{i}}}
\end{equation}

Where $d_{i}$ is the given document, $n_{w},d_{i}$ is the frequence of a word appears in this document.

To predict the class of a document, we only need to calculate $P(c_{+}|d_{i})-P(c_{-}|d_{i})$. If the difference is lager than 0, the document should be predict as positive polarity:

\begin{equation}
    \begin{split}
        P(c_{+}|d_{i}) - P(c_{+}|d_{i}) = \frac{P(c_{+})\prod_{w\in d_{i}}P(w|c_{+})^{n_{w},d_{i}}}{\sum_{r=1}^{C} P(c_{r}) \prod_{w\in d_{i}}P(w|c_{r})^{n_{w},d_{i}}} - \\  
        \frac{P(c_{-})\prod_{w\in d_{i}}P(w|c_{-})^{n_{w},d_{i}}}{\sum_{r=1}^{C} P(c_{r}) \prod_{w\in d_{i}}P(w|c_{r})^{n_{w},d_{i}}} 
    \end{split}
\end{equation}

As we only need to know whether $P(c_{+}|d_{i})-P(c_{-}|d_{i})$ is lager that 0, so the formula could be simplify to:

\begin{equation}
    \begin{split}
P(c_{+}|d_{i}) - P(c_{+}|d_{i}) = 
        P(c_{+})\prod_{w\in d_{i}}P(w|c_{+})^{n_{w},d_{i}} -  
        \\ P(c_{-})\prod_{w\in d_{i}}P(w|c_{-})^{n_{w},d_{i}}
    \end{split}
\end{equation}

\subsection{Discover Words with Sentiment Polarity}

Ideally, if we know the $P(c_{+})$, $P(c_{-})$ and $P(w|c_{j})$ of all words, we can predict the sentiment polarity for all documents. However, above three key components are different in different domains. LSC \cite{chen2015lifelong} proposed a possible solution to calculate $P(w|c_{j})$, but it uses all words which has high risk to be overfitting. As we known, not all words have sentimental polarity like "a", "one" and etc. while some words always have polarity like "good", "hate", "excellent" and so on. In addition, some words only have sentiment polarity in specific domains. For example, "tough" in reviews of the diamond indicates that the diamond have a good quality while it means hard to chew in the domain of food. Hence, in order to achieve the goal of the lifelong learning. We need to find the words always have sentiment polarity and be careful for those words only shows polarity in specific domains. 

\section{Lifelong Semi-supervised Learning for Sentiment Classification}

Although LSC \cite{chen2015lifelong} considered the difference among domains, it still is a typical supervised learning approach.In this paper, we proposed to learn as two stages:
\begin{enumerate}
    \item Initial Learning Stage: to explore a basic set of sentiment words. After that, the model should be able to basically classify a new domain with a good performance.
    \item Self-study Stage: Use the knowledge accumulated from the initial stage to handle new domains, also fine-tune and consolidate the knowledge generated from the initial learning stage. 
\end{enumerate}

\subsection{Initial Learning Stage}
In this stage, we need to train the model to remember some sentiment polarity words. This requires us to find the words with sentiment polarity in each domain. We need to answer two questions here:

\begin{enumerate}
\item How to determine the polarity of a word?
\item How much domains do we need for the initial learning stage?
\end{enumerate}

For the first question, we need to find which words mainly show in the positive or negative documents. This means for a word $w$ with positive polarity, $P(+|w) >> P(-|w)$ or $P(+|w) >> P(+)$. In this paper, we will use $O(w) = P(+|w)/P(+)$ to represent the polarity. This because that the $P(c_{j}|w)/P(w)$ is easy to extend into the multi-classes classification problems. According to the Bayesian formula,  $P(+|w)/P(+) = P(w|+)/P(w)$. 

\subsection{Self-study Stage}

In this stage, our main task is to explore which words have polarity. We will mainly use these words to predict the new domains and assign the pseudo-labels to them. With the pseudo labels, we are able to discover the new words with polarity. Following is the the procedure for self-study:

\begin{enumerate}
\item Using the sentiment words accumulated from the previous tasks to predict a new domain, then assign the prediction results as the pseudo labels.
\item Using the reviews and pseudo labels of above new domain as new training data to run Na\"ive model.
\item Update the sentiment words knowledge base.
\end{enumerate}

\section{Experiment}
\begin{table*}

  \label{tab:commands}
   \begin{tabular}{|l*{10}{|c}|}
\hline
\diagbox{Datasets}{F1 Score}{Percentage}  & 100\% & 90\% & 80\% & 70\% & 60\% & 50\% & 40\% & 30\% & 20\% & 10\% \\
\hline
AlarmClock & 0.8082 & 0.8082 & 0.8082 & 0.8082 & 0.8082 & 0.8082 & 0.8082 & \textbf{0.8082} & 0.274 & 0.2333\\ 
\hline
Baby & 0.6564 & 0.6564 & 0.6564 & 0.6564 & 0.6564 & 0.6564 & 0.6564 & \textbf{0.6564} & 0.1759 & 0.1408\\ 
\hline
Bag & 0.6811 & 0.6811 & 0.6811 & 0.6811 & 0.6811 & 0.6811 & 0.6811 & \textbf{0.6811} & 0.3559 & 0.1056\\ 
\hline
CableModem & 0.6064 & 0.6064 & 0.6064 & 0.6064 & 0.6064 & 0.6064 & 0.6064 & \textbf{0.6064} & 0.2195 & 0.1105\\ 
\hline
Dumbbell & 0.6346 & 0.6346 & 0.6346 & 0.6346 & 0.6346 & 0.6346 & \textbf{0.6346} & 0.6602 & 0.1589 & 0.1383\\ 
\hline
Flashlight & 0.5876 & 0.5876 & 0.5876 & 0.5876 & 0.5876 & 0.5876 & 0.5876 & \textbf{0.5921} & 0.3278 & 0.1036\\ 
\hline
Gloves & 0.6131 & 0.6131 & 0.6131 & 0.6131 & 0.6131 & 0.6131 & 0.6131 & \textbf{0.6131} & 0.3205 & 0.1206\\ 
\hline
GPS & 0.6814 & 0.6814 & 0.6814 & 0.6814 & 0.6814 & 0.6814 & 0.6814 & \textbf{0.6814} & 0.2838 & 0.1629\\ 
\hline
GraphicsCard & 0.5775 & 0.5775 & 0.5775 & 0.5775 & 0.5775 & 0.5775 & 0.5775 & \textbf{0.5775} & 0.2776 & 0.1271\\ 
\hline
Headphone & 0.6578 & 0.6578 & 0.6578 & 0.6578 & 0.6578 & 0.6578 & 0.6578 & \textbf{0.6578} & 0.268 & 0.1745\\ 
\hline
HomeTheaterSystem & 0.8394 & 0.8394 & 0.8394 & 0.8394 & 0.8394 & 0.8394 & 0.8394 & \textbf{0.8394} & 0.2404 & 0.2238\\ 
\hline
Jewelry & 0.604 & 0.604 & 0.604 & 0.604 & 0.604 & 0.604 & 0.604 & \textbf{0.604} & 0.3371 & 0.1088\\ 
\hline
Keyboard & 0.653 & 0.653 & 0.653 & 0.653 & 0.653 & 0.653 & 0.653 & \textbf{0.653} & 0.2117 & 0.1841\\ 
\hline
MagazineSubscriptions & 0.8042 & 0.8042 & 0.8042 & 0.8042 & 0.8042 & 0.8042 & 0.8042 & 0.8042 & \textbf{0.8049} & 0.2115\\ 
\hline
MoviesTV & 0.5843 & 0.5843 & 0.5843 & 0.5843 & 0.5843 & 0.5843 & 0.5843 & 0.5843 & \textbf{0.606} & 0.0976\\ 
\hline
Projector & 0.7387 & 0.7387 & 0.7387 & 0.7387 & 0.7387 & 0.7387 & 0.7387 & \textbf{0.7387} & 0.1814 & 0.168\\ 
\hline
RiceCooker & 0.7656 & 0.7656 & 0.7656 & 0.7656 & 0.7656 & 0.7656 & \textbf{0.7656} & 0.7739 & 0.1683 & 0.1566\\ 
\hline
Sandal & 0.5987 & 0.5987 & 0.5987 & 0.5987 & 0.5987 & 0.5987 & 0.5987 & \textbf{0.5987} & 0.3501 & 0.1077\\ 
\hline
Vacuum & 0.7362 & 0.7362 & 0.7362 & 0.7362 & 0.7362 & 0.7362 & 0.7362 & \textbf{0.7362} & 0.2155 & 0.1807\\ 
\hline
VideoGames & 0.6835 & 0.6835 & 0.6835 & 0.6835 & 0.6835 & 0.6835 & 0.6835 & \textbf{0.6835} & 0.4514 & 0.173\\ 
\hline
Average & 0.6756 & 0.6756 & 0.6756 & 0.6756 & 0.6756 & 0.6756 & 0.6756 & \textbf{0.6775} & 0.3114 & 0.1514\\ 
\hline

\end{tabular}
\caption{F1 Score of Na\"ive Bayesian Classifiers under Decreasing Word Usage Percentage }
\end{table*}

\subsection{Datasets}

In the experiment, we use the same datasets as LSC \cite{chen2015lifelong} used. It contains the reviews from 20 domains crawled from the Amazon.com and each domain has 1,000 reviews (the distribution of positive and negative reviews is imbalanced). 

\subsection{Word Polarity Analysis}

To answer the first question for the initial learning stage, we need to know which words exactly influence the sentiment classification. Firstly, we calculate $P(w|+)$ and $P(w|-)$ for each words. Then, we define the polarity degree by $O(w) = P(w|+)/P(w)$. Finally, we only choose a specific percentage words to predict and see whether the performance decreases. In addition, we also only consider the words that at least show over average 5 times in per domain. This because that we did not delete the symbols and numbers in the data, and these characters may be noise in the training data.

We firstly sorted the words or symbols (no data pre-processing to the corpus in this paper) by the polarity $O(w)$ and then choose a specific percentage words or symbols from the whole words to only 10\%. From Table 1 we can see that using no less than 30\% can obtains the best average result. This means that the most of words and symbols do not have obvious sentiment orientation. 

Hence, we will only keep 30\% of words for Na\"ive Bayes model and even get better f1 score. Although the performance decrease on a single domain, the better global performance can achieve with only the sentiment words. 

\subsection{Requirement for the Initial Learning}
For the second question of the initial learning stage, the answer depends on the tasks. In the practice, all of the labeled data definitely need to be used for training. The only question should be conceded is that how much labeled data can meet the minimum requirement. For this sentiment classification task, one domain is absolutely insufficient. Based on the experiment result, the initial learning stage at least needs two domains, and can achieve much better performance with three domains. Increase more domains will not significant influence the performance. Hence, three domains are enough for this task. For different tasks, two labeled domains are necessary. More labeled domains are suggested to continue collect until the performance on the new domains tends to steady.

\subsection{Self-study Learning}

In the self-study learning stage, the learning is designed under the unsupervised learning setting. In this stage, there is any labeled data. Instead of that, we uses the model generate from the initial learning stage to predict each domain and assign the pseudo labels to them. After that, the model will regard the pseudo labels as the real labels and continue the training on the new domain. With this method, self-study learning stage can learn new domains well without any labeled data.

Table 2 is the F1 score of three models on 17 domains. The first three domains was used for the initial learning stage. And we use the Macro-F1 score because the datasets are imbalanced and it can prove our performance on the minor classes. We compared our model (Semi-Unsupervised Learning, SU-LML for short) with Na\"ive Bayes model which only trained on the first three (source) domains (NB-S) and Na\"ive Bayes model trained on each domain with labels by 5-fold cross validation (NB-T). We can see that our approach is significantly better than other two approaches. It even perform better than the NB-T, a typically supervised learning. The figure 2 shows the result more clearly. The comparisons to LSC and neural based lifelong learning \cite{lv2019sentiment} are not going to show here, because firstly their codes are still unavailable and secondly their approaches are totally supervised learning. 

\begin{table}

  \label{tab:commands}
   \begin{tabular}{|l*{10}{|c}|}
\hline
\diagbox{Datasets}{F1 Score}{Model}  & NB-S & NB-T & SU-LML\\
\hline
CableModem &  0.4774 & 0.6633 & \textbf{0.8694}\\
\hline
Dumbbell &  0.6539 & 0.764 & \textbf{0.8748}\\
\hline
Flashlight &  0.6536 & 0.6251 & \textbf{0.8259}\\
\hline
Gloves &  0.5973 & 0.6943 & \textbf{0.785}\\
\hline
GPS &  0.6447 & 0.7465 & \textbf{0.9121}\\
\hline
GraphicsCard &  0.4797 & 0.7346 & \textbf{0.8768}\\
\hline
Headphone &  0.5938 & 0.7356 & \textbf{0.8858}\\
\hline
HomeTheaterSystem &  0.6242 & 0.8611 & \textbf{0.9236}\\
\hline
Jewelry &  0.6927 & 0.7088 & \textbf{0.7599}\\
\hline
Keyboard &  0.6905 & 0.7289 & \textbf{0.8707}\\
\hline
MagazineSubscriptions &  0.6284 & 0.8056 & \textbf{0.8932}\\
\hline
MoviesTV &  0.4991 & 0.6785 & \textbf{0.8381}\\
\hline
Projector &  0.6565 & 0.7525 & \textbf{0.8575}\\
\hline
RiceCooker &  0.6833 & 0.8027 & \textbf{0.8475}\\
\hline
Sandal &  0.6972 & 0.6904 & \textbf{0.8059}\\
\hline
Vacuum &  0.7728 & 0.8 & \textbf{0.8992}\\
\hline
VideoGames &  0.5665 & 0.7564 & \textbf{0.9068}\\
\hline
Average &  0.6242 & 0.7381 & \textbf{0.8607}\\
\hline

\end{tabular}
\caption{F1 Score for NB-S, NB-T, SU-LML}
\end{table}

\begin{figure}[h]
  \centering
  \includegraphics[width=\linewidth]{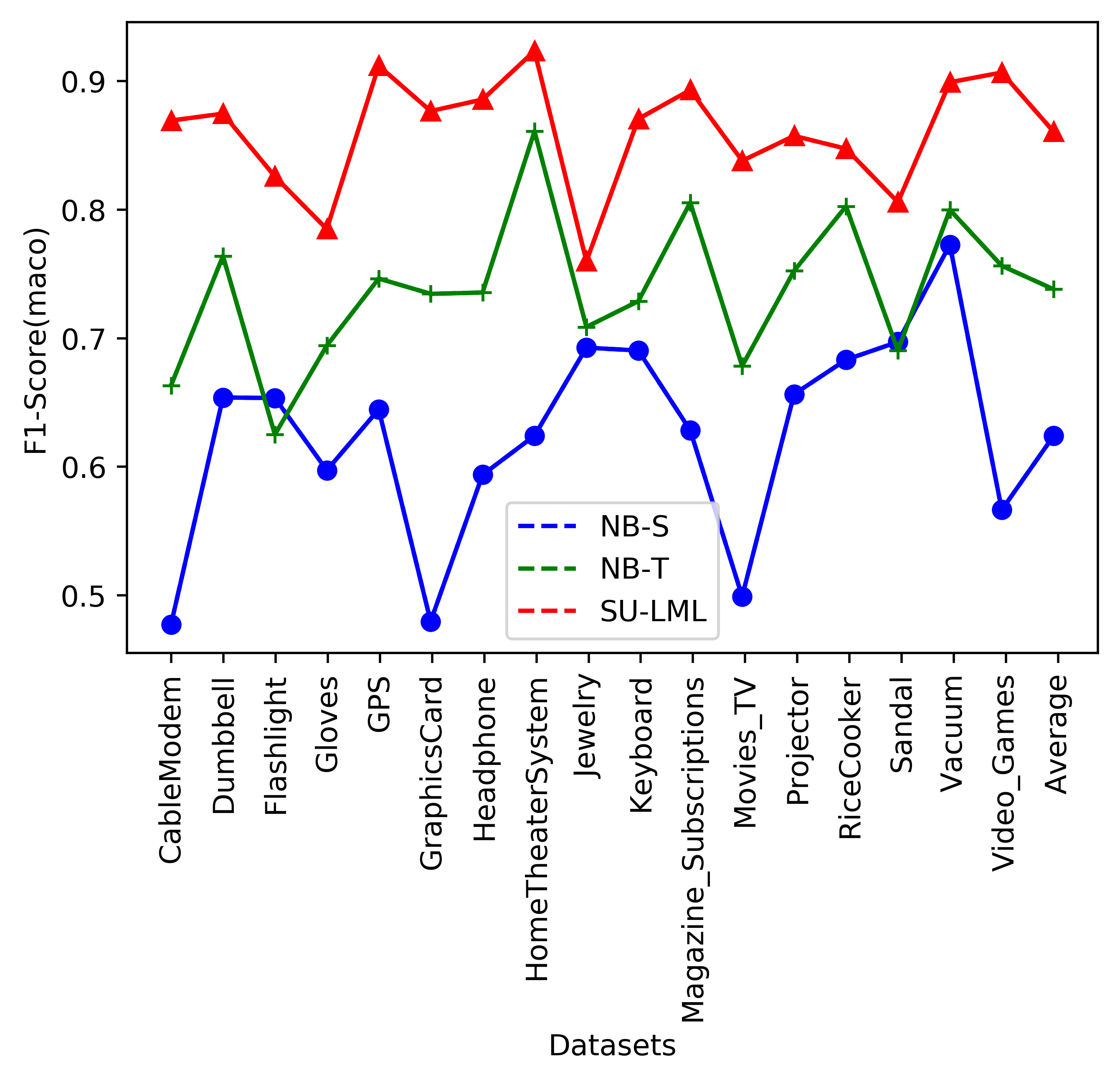}
  \caption{F1 Score in Self-Study Stage}
  \Description{The 1907 Franklin Model D roadster.}
\end{figure}

\begin{table}[]
    \centering
    \begin{tabular}{c|c}
    \hline
         Word  &  Degree for Negative Sentiment \\ 
         \hline
        refund  &  32.99921813917123 \\ 
garbage  &  32.994266353922335 \\ 
junk  &  32.985405264529575 \\ 
waste  &  32.984102163148286 \\ 
worst  &  32.97185301016418 \\ 
rma  &  32.96846494657285 \\ 
poorly  &  32.96194943966641 \\ 
terrible  &  32.95569455303623 \\ 
disappointed  &  32.949960906958566 \\ 
trash  &  32.948918425853535 \\ 
useless  &  32.94683346364347 \\ 
worthless  &  32.94057857701329 \\ 
awful  &  32.92520198071411 \\ 
defective  &  32.917904612978894 \\ 
return  &  32.913734688558776 \\ 
exchange  &  32.908001042481104 \\ 
respond  &  32.90487359916601 \\ 
poor  &  32.90409173833724 \\ 
disappointment  &  32.90278863695596 \\ 
crap  &  32.89653375032577 \\ 
\hline
    \end{tabular}
    \caption{Top 20 Words with Negative Sentiment}
    \label{tab:my_label}
\end{table}

\subsection{Knowledge Generated during Learning}
In this paper, we done one more important things is that we discovered which words have sentiment polarity. If a word was regarded with sentiment polarity, we increase the polarity score of it with one. In addition, we will plus an additional score from 0 to 1 to 1 based on the $O(w)$ rank. From table 3, we can see that most top words with negative emotion and most of them make sense.  

\section{Conclusion and Outlook}
We proposed a semi-unsupervised lifelong sentiment classification approach in this paper. It can accumulate knowledge from the previous learning and turn to self-study. A very few labeled data required in our approach so it is very suitable for the industry scenario. The performance of it even exceeds the supervised learning, which shows that the knowledge reusing of the lifelong learning is useful. 

Although we only show two classes classification here, but the ideal is also suitable for the multi-classes classification. All text classification can use this approach, not only sentiment classification. Our model classify documents by the knowledge of the sentiment polarity of the words, which uses the same approach of we human being. We shows that to focus the goal behind the learning tasks is more meaningful than just to find a solution. Understanding the words is much important than solve a sentiment classification task. We should learn the knowledge and skills for all tasks rather than a solution for a single task. 

\section{Acknowledgments}

This research is supported by the Research Institute of Big Data Analytics, Xi’an Jiaotong – Liverpool University and the CERNET Innovation Project under Grant NGII20161010.

%
\bibliographystyle{ACM-Reference-Format}
\bibliography{reference.bib}

%

\end{document}